\documentclass[final,3p,times]{elsarticle}

\usepackage{amssymb}
\usepackage{amsmath}

\usepackage{graphicx}
\usepackage{xcolor}
\usepackage{adjustbox}
\usepackage{multirow}
\usepackage{array}
\usepackage{threeparttable}
\PassOptionsToPackage{hyphens}{url}\usepackage{hyperref}
\usepackage{comment}

\journal{Computers in Biology and Medicine}

\begin{document}

\begin{frontmatter}



\title{Empirical investigation of multi-source cross-validation in clinical ECG classification} 


\author[utu]{Tuija Leinonen}
\author[leeds]{David Wong}
\author[utu]{Antti Vasankari}
\author[leeds_crd]{Ali Wahab}
\author[leeds_crd]{Ramesh Nadarajah}
\author[utu]{Matti Kaisti}
\author[utu]{Antti Airola\corref{cor1}} \ead{ajairo@utu.fi}
\cortext[cor1]{Corresponding author.}
\affiliation[utu]{organization={Department of Computing, University of Turku},
            country={Finland}}

\affiliation[leeds]{organization={Leeds Institute of Health Sciences, University of Leeds},
            country={UK}}

\affiliation[leeds_crd]{organization={Institute of Cardiovascular and Metabolic Medicine, University of Leeds},
            country={UK}}

\begin{abstract}
Traditionally, machine learning-based clinical prediction models have been trained and evaluated on patient data from a single source, such as a hospital. Cross-validation methods can be used to estimate the accuracy of such models on new patients originating from the same source, by repeated random splitting of the data. However, such estimates tend to be highly overoptimistic when compared to accuracy obtained from deploying models to sources not represented in the dataset, such as a new hospital. The increasing availability of multi-source medical datasets provides new opportunities for obtaining more comprehensive and realistic evaluations of expected accuracy through source-level cross-validation designs.

In this study, we present a systematic empirical evaluation of standard K-fold cross-validation and leave-source-out cross-validation methods in a multi-source setting. We consider the task of electrocardiogram based cardiovascular disease classification, combining and harmonizing the openly available PhysioNet/CinC Challenge 2021 and the Shandong Provincial Hospital datasets for our study.

Our results show that K-fold cross-validation, both on single-source and multi-source data, systemically overestimates prediction performance when the end goal is to generalize to new sources. Leave-source-out cross-validation provides more reliable performance estimates, having close to zero bias though larger variability. The evaluation highlights the dangers of obtaining misleading cross-validation results on medical data and demonstrates how these issues can be mitigated when having access to multi-source data.
\end{abstract}

\begin{keyword}

Performance evaluation \sep Multi-source cross-validation \sep Electrocardiogram classification \sep Cardiovascular diseases \sep Deep learning 

\end{keyword}

\end{frontmatter}



\section{Introduction}

The electrocardiogram (ECG) is a key tool for diagnosing and monitoring cardiovascular diseases (CVD). Advancements in machine learning have opened up the potential to create new predictive models for disease diagnosis based on ECG. Traditionally, such models have been developed using data from a single source such as a hospital \cite{hannun2019, Kiranyaz2016, Celin2018, Liu2023}. In such contexts, standard holdout methods where data is split to training, validation and test sets, or cross-validation methods are commonly used to estimate the prediction performance of the developed models. Whilst these methods can reliably estimate the expected performance for future patients originating from the same source, the evaluations have been shown to be optimistically biased if a model is used to make predictions on data originating from new source \cite{Merdjanovska2021}. The reason for this bias may include differences in underlying patient populations, variations in labeling practices ranging from manual to computer-assisted methods, differing diagnostic practices for CVDs across different countries \cite{Rajpurkar2022, han2020survey, zhao_repo} and collection of imbalanced or unrepresentative training data \cite{norori2021addressing}.

Increased availability of multi-source data that include patients collected across multiple sites offers new possibilities for more reliable performance estimation. Even in this context, standard holdout and cross-validation methods based on random splitting of the combined data can be expected to be overoptimistic when deploying models to sources not represented in the dataset. In this paper, we present an empirical study that explores the extent this issue can be mitigated by using a leave-source-out cross-validation \cite{geras2013multiple} approach. The method simulates the setting in which a model trained on data combined from a number of sources is applied to data originating from a new source that is not part of the training data. This approach simulates the expected prediction performance when deploying a model to a new source. The main contributions of this study are as follows:

\begin{itemize}
\item We combine and harmonize the openly available PhysioNet/CinC challenge 2021 \cite{reyna2021will, cinc2021_longer, PhysioNet} and the Shandong Provincial Hospital \cite{shandong} datasets, creating a large multi-source resource for testing the generalization of ECG-based diagnostic classifiers for CVDs.
\item We empirically evaluate how reliable standard K-fold cross-validation estimates are when a classifier trained and validated on data from a single source is later applied for making predictions for patients from a different source, e.g. hospital.
\item We further evaluate whether leave-source-out cross-validation provides more reliable estimates than the standard K-fold cross-validation regarding the generalization of the model to a new source, in settings where data from multiple sources are available for both model training and validation. 
\item Finally, we introduce and apply a classification based heuristic for detecting systematic differences between the source distributions.
\end{itemize}

In this study, we employ four classification algorithms: state-of-the-art deep learning models, including residual networks \cite{zhao_repo} and transformers [15], alongside logistic regression and XGBoost based baseline methods. To evaluate reliability of different cross-validation approaches, we introduce a comprehensive metric that analyzes mean error (bias), standard deviation, and mean squared error of their estimates. All datasets and code used for these experiments are publicly available at \url{https://github.com/UTU-Health-Research/dl-ecg-classifier}. Overall, our work presents the most comprehensive evaluation to date of the source-level cross-validation approach on clinical ECG data.

In Section~\ref{sec:rel_work} we cover related work on estimating generalization of biomedical classifiers. Section~\ref{sec:mat_met} covers the datasets and their harmonization, machine learning methodologies, and experimental setup. In Section~\ref{sec:results} we present the experimental results, in Section~\ref{sec:discussion} the discussion, and Section~\ref{sec:conclusion} concludes the paper.

\section{Related work}\label{sec:rel_work}

Many studies have examined challenges in deployment of AI models in healthcare, and shown that performance metrics reported in the literature tend to be overoptimistic. Factors contributing to these challenges include limited data access \cite{Jiang230,kulkarni2021}, differing diagnostic practices and ambiguous ground truth \cite{Rajpurkar2022, han2020survey,kulkarni2021}, biased data \cite{norori2021addressing} and differences between the distribution of data in training and test sites \cite{kulkarni2021,padovano2022hidden,white2018}.

A recent study by Kapoor et al. \cite{kapoor2023} exposed the pitfalls of poorly chosen evaluation approaches in data analysis, leading to over-optimistic performance claims due to data leakage across hundreds of studies. Such issues arise, in part, from the lack of a truly separate test set and the non-independence of training and testing samples. This problem is particularly pronounced when dealing with real-world sources like medical data.

Several studies have demonstrated that cross-validation (CV) results tend to be optimistically biased if the goal is to evaluate generalization in a new independent data source. For example, Padovano et al. \cite{padovano2022hidden} presented drastically different results in obstructive sleep apnea classification when using an external test set compared to standard 10-fold CV. White et al. \cite{white2018} illustrated how the K-fold CV may lead to performance overestimation due to random splitting of samples from a single trial into training and testing sets, rather than using all the samples from the given trial for either training or testing. In related research on clinical prediction models, use of external validation sets has been proposed as means to obtain more reliable estimates about true generalization to new patient populations \cite{bleeker2003external}.

Rakotomalala et al. \cite{rakotomalala2006accuracy} considered the use of CV for multi-source data, using hospital wards as an example. They empirically demonstrated, that if the goal is to estimate generalization performance with respect to new sources, standard K-fold CV produces overoptimistic results. They proposed an alternative approach where CV folds are sampled on the level of data sources, rather than on the level of individual patients. Geras et al. \cite{geras2013multiple} called such approach multiple-source cross-validation, and theoretically showed that it produces an unbiased estimate of the expected prediction performance on data from a new source. Multi-source evaluation is closely related to a common scenario in medical data analysis, where data is grouped by subjects, each contributing multiple data points, such as medical scans or biosignal measurements. In these cases, subject-level CV methods, such as leave-subject-out \cite{esterman2010avoiding,saeb2017need}, are frequently used to ensure robust model performance across different individuals. Multi-source CV extends this approach to evaluate generalization across entire data sources, such as hospitals, rather than focusing solely on the individual subject level.

Multi-source CV has been applied in a number of previous studies within the medical field. Knight et al. \cite{knight2018voxel} empirically evaluated leave-one-source-out CV for Magnetic Resonance Imaging where data was gathered from different scanners. Han et al. \cite{han2021towards} assessed their two models for 12-lead ECG classification, calling multiple-source CV as leave-one-dataset-out CV. Tabe-Bordbar et al. \cite{borbdar2018} compared the clustering-based CV with the standard CV on the task of gene expression prediction and illustrated how the standard CV produced over-optimistic performance estimates. McWilliams et al. \cite{mcwilliams2019towards} implemented the CV approach with data from two sources for the task of intensive care discharge prediction. Compared to these previous works, our study includes both a greater number of data sources, a greater volume of data, i.e. individuals, and more systematic evaluation protocol for assessing the bias, variance and overall mean squared error of the compared cross-validation methods. 

\section{Materials and Methods}\label{sec:mat_met}

\subsection{Data sources}

In our study, we used public training data from the George B. Moody PhysioNet/CinC challenge 2021 (CinC)\cite{reyna2021will, cinc2021_longer, PhysioNet} and a separate publicly available database from the Shandong Provincial Hospital (SPH) \cite{shandong}. Both data contain 12-lead ECGs with patient and recording information and diagnostics labels. A general overview of the CinC data as well as the SPH data is presented in Table \ref{table:databases_overview}. Considering our research objectives involving multi-source data, the data is treated as five separate data sources, four of which are part of the PhysioNet/CinC challenge 2021 data. Some of the data and diagnostic labels were removed from this study, as detailed below, resulting in a total of 103,438 ECGs (CinC: 80,164 and SPH: 23,274). The following descriptions of the datasets only consider this subset of the data, together forming the final dataset for our study.

\begin{table}[t]
\centering
\caption{Overview of the five data sources in the study. All sources, except for SPH \cite{shandong}, are part of the PhysioNet/CinC challenge 2021 training set \cite{reyna2021will, cinc2021_longer, PhysioNet}. Numbers of patients and published ECGs correspond to the published numbers about the challenge training set: PhysioNet divided G12EC and CPSC (not CPSC-Extra) into training, validation and test and published only the training set. The final dataset for our study contained 103,438 recordings, including only ones with particular diagnoses.} \label{table:databases_overview}
\begin{adjustbox}{width=1\textwidth,center=\textwidth}

\bgroup
\def\arraystretch{1.6}
\begin{tabular}{lllcccccc}
\hline
\textbf{Sources}   & \textbf{Countries}   & \textbf{Locations}   & \textbf{\begin{tabular}[c]{c}\vspace{-0.5em}Total \\ patients (n)\end{tabular}} & \textbf{\begin{tabular}[c]{c}\vspace{-0.5em} Total \\ ECGs (n) \end{tabular}} & \textbf{\begin{tabular}[c]{c}\vspace{-0.5em}Included \\ ECGs (n)\end{tabular}} & \textbf{\begin{tabular}[c]{@{}c@{}}\vspace{-0.5em}Sampling \\ frequency\end{tabular}} & \textbf{\begin{tabular}[c]{c}\vspace{-0.5em}Length\\ (s)\end{tabular}} & \textbf{\begin{tabular}[c]{c}\vspace{-0.5em}Labeling\\ standard\end{tabular}}\\  \hline
\begin{tabular}[c]{@{}l@{}}\vspace{-0.5em}Chapman-Shaoxing\\ and Ningbo\end{tabular} & China  & \begin{tabular}[c]{@{}l@{}}\vspace{-0.5em}Shaoxing People's Hospital\\ Ningbo First Hospital\end{tabular}     & 45,152      & 45,152 & 43,814  & 500   & 10   & SNOMED   \\
\begin{tabular}[c]{@{}l@{}}\vspace{-0.5em}CPSC and\\ CPSC-Extra\end{tabular} & China  & 11 unnamed hospitals   & Unknown   & 10,330 & 6,110  & 500   & 6-144   & SNOMED  \\
G12EC  & USA    & Emory University Hospital  & 15,738   & 10,344 & 8,892  &  500  & 5-10   & SNOMED \\ 
PTB and PTB-XL  & \begin{tabular}[c]{@{}l@{}}\vspace{-0.5em}Germany and other\\ European countries\end{tabular} & \begin{tabular}[c]{@{}l@{}}\vspace{-0.5em}University Clinic Benjamin Franklin\\ Physkalisch Technische Bundesantalt\end{tabular} & 19,147     & 22,353 & 21,348   & 500, 1000    & 10-120              & SNOMED   \\ 
SPH      & China   & Shandong Provincial Hospital   & 24,666    & 25,770 & 23,274 & 500  & 10-60 & AHA \\ \hline
\end{tabular}
\egroup
\end{adjustbox}
\end{table}

\subsubsection{George B. Moody PhysioNet/CinC Challenge 2021 Dataset}

The CinC data, including only recordings with labels of our interest, consist of 80,164 12-lead ECGs from multiple databases. Among these databases, some originate from the same location, which we addressed by integrating them into a unified dataset based on their shared origin. For instance, PhysioNet derived two separate databases from the China Physiological Signal Challenge 2018 (CPSC), the CPSC database and the unused CPSC-Extra database \cite{liu2018open, cinc2021_longer}. We treated them as a unified data source, naming it \textit{the CPSC and CPSC-Extra source}. Similar merging was applied to the Physikalisch-Technische Bundesanstalt (PTB) and PTB-XL databases \cite{ptb, ptb-xl} and the Chapman-Shaoxing and Ningbo databases \cite{chapman, ningbo}. Consequently, we considered the CinC data as four distinct data sources, the CPSC and CPSC-Extra source, the PTB and PTB-XL source, the Chapman-Shaoxing and Ningbo source and the Georgia 12-lead ECG (G12EC) source \cite{g12ec}.

We decided to exclude the St Petersburg INCART 12-lead Arrhythmia database \cite{incart} from the final dataset due to substantial dissimilarities in size and duration compared to the other sources. The database contains 74 ECG recordings, each lasting 30 minutes, while the others have many more ECG recordings (several thousand) of much shorter duration (varying from 5 seconds to 144 seconds). Out of the 74 recordings in the INCART database, only 38 ECGs are labeled with the diagnoses that were used in the evaluation of the PhysioNet/CinC challenge 2021, and just 21 ECGs include the labels that we selected for our study.

\begin{figure}
    \centering
    \includegraphics[width=0.8\linewidth]{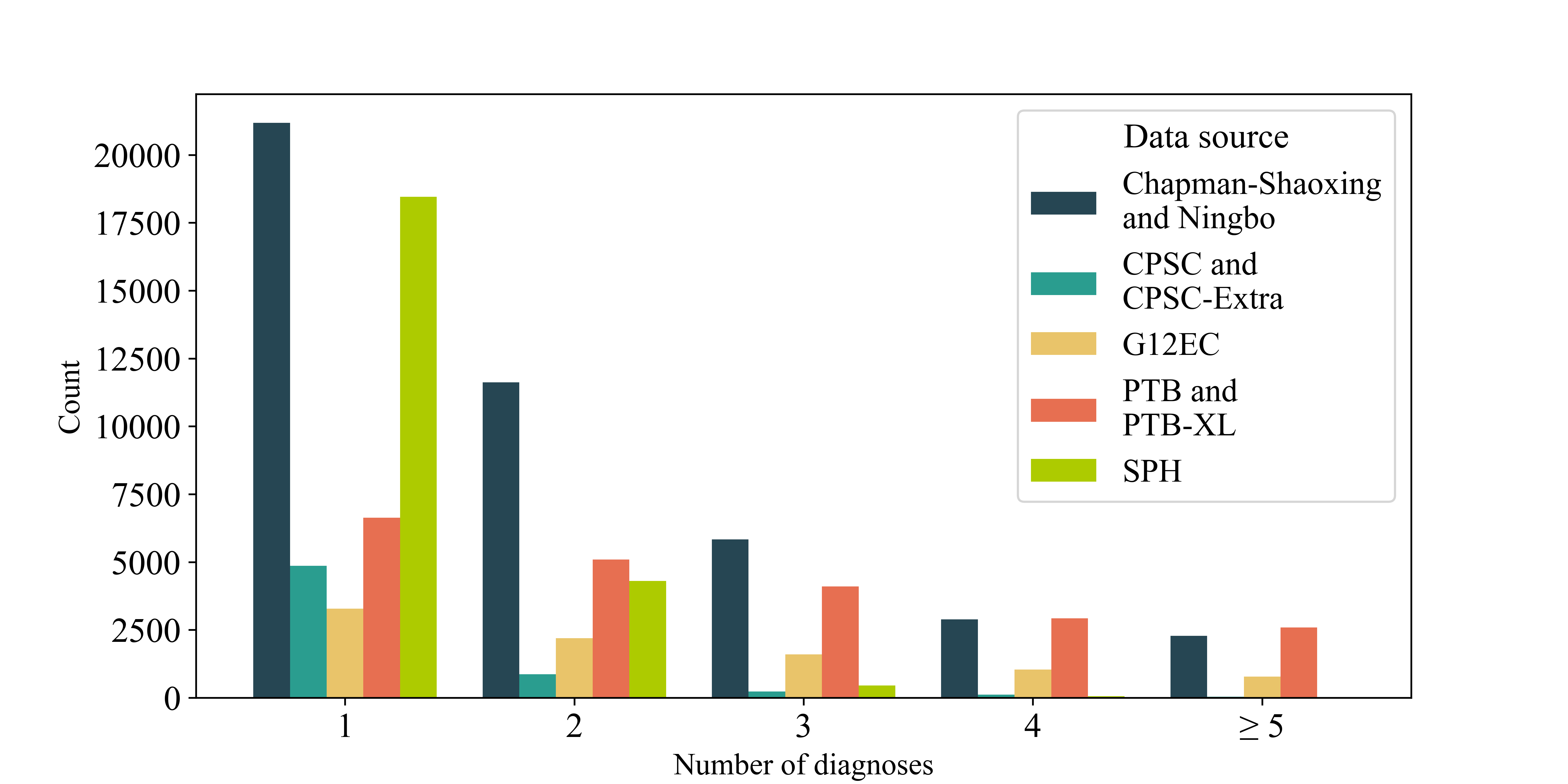}
    \caption{Number of diagnoses per patient per data source. The last bar summarizes the range from 5 to 12.}
    \label{fig:label_occur}
\end{figure}

 The majority of the ECGs in the CinC data are sampled at 500 Hz, but a small number are sampled at 1000 Hz in the PTB and PTB-XL source. The duration of the recordings spans from 5 seconds to nearly 2.5 minutes. Most of the recordings include age and sex as demographic information and more detailed lead and recording information. Table \ref{table:demo_info} describes the age and sex attributes per data source. Whilst the majority of the ECGs include one (44.9 \%), two (24.7 \%) or three (14.7 \%) labels, the total number of labels can vary from 1 to 12, depending on the data source (Figure \ref{fig:label_occur}). Table~\ref{table:databases_overview} indicates that in the PTB and PTB-XL source, the number of patients is lower than the number of included ECGs in the study, which suggests the presence of multiple recordings for some patients. However, due to the lack of the patient identifiers within the CinC data, we were unable to examine the potential for duplicate recordings per patient.

\subsubsection{Shandong Provincial Hospital Database}

The SPH database, released in 2022, was collected at Shandong Provincial Hospital between 2019 and 2020. We treated it as an independent data source in addition to the four sources that form the PhysioNet/CinC challenge 2021 data. Once the data source was filtered to contain only the particular diagnoses chosen for our study, the SPH source contains 23,274 annotated 12-lead ECGs. Of these, 59.7 \% of the ECGs are labeled as "normal ECGs" indicating they do not include any diagnosed abnormalities. As shown in Figure~\ref{fig:label_occur}, 79.3 \% of all the recordings have a single label. 18.5 \% have two labels, 1.9 \% have three labels, and 0.3 \% are assigned either four or five labels.

The ECGs are sampled at 500 Hz with a duration spanning between 10 and 60 seconds. For each ECG, age and sex information is provided (Table \ref{table:demo_info}) as well as the recording date, the number of samples, and the patient ID. The portion of the SPH source extracted for our study contains ECGs from 22,334 patients, some of whom have multiple ECGs, albeit the majority of them (21,540) have only a single recording. We compared the shared metadata of the ECGs per patient and noticed that multiple ECGs shared identical metadata, including diagnoses, patient ID, age, sex, the number of samples, and the date the ECG was recorded. In total, 144 ECGs had duplicates based on the metadata alone, and of these, 132 (0.13 \% of the final dataset) had identical ECG recording data.

Instead of SNOMED CT codes, the SPH source follows the American Heart Association (AHA), for ECG labeling, as recommended by the AHA, the American College of Cardiology, and the Heart Rhythm Society. The labels are provided as primary statements and optional additional modifiers. The primary statements are single labels conveying clinical meaning as is, and the modifiers serve as adjectives to further refine the meaning of the core statement \cite{aha_standard_2007}. 

\begin{table}[]
\centering
\caption{Demographic information per data source} \label{table:demo_info}

\begin{adjustbox}{width=0.8\textwidth,center=\textwidth}
\bgroup
\def\arraystretch{1.6}

\begin{tabular}{lccccc}
\hline
\multirow{2}{*}{\textbf{Source}} & \multicolumn{4}{c}{\textbf{Age}}  & \multicolumn{1}{c}{\textbf{Sex}}\\ 
\cline{2-6} 
\multicolumn{1}{c}{}  & \multicolumn{1}{l}{\textbf{Mean ($\pm$ SD)}} & \multicolumn{1}{l}{\textbf{Min}} & \multicolumn{1}{l}{\textbf{Max}} & \multicolumn{1}{l}{\textbf{U (n)}} & \textbf{\begin{tabular}[c]{@{}c@{}}\vspace{-0.5em}Ratio \% \\ (Male/Female)\end{tabular}}\\ \hline
Chapman-Shaoxing and Ningbo       & 58.21 ($\pm$ 19.62) & 0 & 89 & 44  & 56/43   \\
CPSC and CPSC-Extra               & 62.69 ($\pm$ 18.67) & 1 & 104 & 8  & 57/43    \\
G12EC                             & 60.82 ($\pm$ 15.51) & 14 & 89 & 65 & 54/46  \\
PTB and PTB-XL                   & 59.57 ($\pm$ 17.01)& 2 & 95 & 92 & 52/48  \\
SPH                                & 49.66 ($\pm$ 15.51)  & 18 & 95 & 0 & 57/43 \\ \hline
\textbf{Final dataset}                      & 57.05 ($\pm$ 18.32) & 0 & 104 & 209 & 56/44 \\ \hline
\multicolumn{6}{l}{\footnotesize U = Undefined $\vert$  The Chapman-Shaoxing and Ningbo source has 14 ECGs that lack the sex value.}
\end{tabular}
\egroup
\end{adjustbox}
\end{table}

\subsubsection{Mapping between SNOMED CT and AHA standards}

The CinC data uses the SNOMED CT standard, whereas the SPH source uses the AHA standard for labeling. We created a label mapping to convert AHA codes to SNOMED CT. The mapping was created based on the provided descriptions of the CinC data \cite{reyna2021will} and the SPH source \cite{shandong} in collaboration with two cardiologists. The mapping is shown in Table \ref{table:databases_diagnoses}.

The mapping has the following nuances. First, in the SPH source, ECGs labeled with a "normal ECG" label were assigned a value of -1 for easier identification and to prevent their removal. Second, we merged all the prolonged PR interval labels to the first degree atrioventricular (AV) block label, as the first degree AV block diagnosis was missing from the SPH source but both were found in the CinC data. Third, as described, the SPH source includes both the primary statements, which can be seen as main diagnostic labels, and the modifiers that refines those statements. For example, atrial fibrillation can be used as a primary statement and it can be further characterized using two modifiers, "with a rapid ventricular response" or "with a slow ventricular response". This results in three distinct diagnostic categories, each containing different ECG recordings. As the SNOMED CT standard does not include such refinements, we merged all the ECGs under atrial fibrillation and its further refinements under "atrial fibrillation". Similar merging was applied to ECGs indicating premature atrial contraction, which initially was refined using five additional modifiers.

After applying the label mapping, ECGs lacking any labels were excluded from the SPH data. This step was taken to remove all recordings that only had diagnostic labels that were excluded from this study. Thus, these recordings were characterized as not being "normal ECGs" and had no diagnostic label outlined in Table \ref{table:databases_diagnoses}. In total, 2,496 ECGs (of 25,770) were removed after the mapping process. 

\subsubsection{Selection of labels}

The labels were chosen following specific criteria. The first criterion was that each label must be included in the PhysioNet/CinC challenge 2021 evaluation, i.e. it should be found among the 30 scored diagnoses. These 30 diagnoses were chosen based on their relevance by the PhysioNet challenge team, which included expert cardiologist guidance. The second criterion required that each label had to be present in a minimum of four distinct data sources and occur at least 50 times in each of them. In other words, it was acceptable if a chosen diagnosis was missing or had a low number of instances in one source, as long as it was well-represented in the remaining sources.

Based on the above mentioned criteria, a total of 16 diagnoses along with the sinus rhythm label were selected for ECG classification. Table \ref{table:databases_diagnoses} shows the diagnoses and their respective occurrences in each data source. The data sources are highly imbalanced, with significant differences in label distributions across each source. A key objective of this study is to simulate real-world conditions where data is aggregated from multiple hospitals and to examine how these inherent disparities impact model generalization and performance evaluation.

 \begin{table}[]
\centering
\caption{Numbers of selected diagnoses within the data sources in our study. The plus sign in the AHA statements is used to joint a primary statement and an additional modifier.} \label{table:databases_diagnoses}
\begin{adjustbox}{width=\textwidth, center=\textwidth}
\begin{threeparttable}
\bgroup
\def\arraystretch{1.8}
\begin{tabular}{lllccccc>{\bfseries}c}
\hline

\textbf{Diagnosis} & \textbf{SNOMED CT} & \textbf{AHA} & \multicolumn{1}{c}{\textbf{\begin{tabular}[c]{c}\vspace{-0.5em}Chapman-Shaoxing\\ and Ningbo\end{tabular}}} & \multicolumn{1}{c}{\textbf{\begin{tabular}[c]{c}\vspace{-0.5em}CPSC and\\ CPSC-Extra\end{tabular}}} & \textbf{G12EC} & \multicolumn{1}{c}{\textbf{\begin{tabular}[c]{c}\vspace{-0.5em}PTB and\\ PTB-XL\end{tabular}}} & \textbf{SPH} & \textbf{Total} \\ \hline
first degree AV block\tnote{1}     & 270492004  & 82 & 1192& 828 & 769 & 1012 & 238  & 4039    \\ 
atrial fibrillation                  & 164889003  & \begin{tabular}[c]{@{}l@{}} \vspace{-0.5em} 50\\ \vspace{-0.5em} 50+346\\ 50+347 \end{tabular}  & 1780  & 1374  & 570  & 1529  & 675 & 5928  \\ 
atrial flutter                       & 164890007 & 51 & 8060  & 54 & 186  & 74   & 99  & 8473  \\ 
\begin{tabular}[c]{@{}l@{}}\vspace{-0.5em}incomplete right\\bundle branch block\end{tabular} & 713426002  & 105  & 246  & 86  & 407 & 1118  & 1259  & 3116 \\ 
\begin{tabular}[c]{@{}l@{}}\vspace{-0.5em}left anterior\\ fascicular block\end{tabular}       & 445118002  & 101 & 380   & 0   & 180  & 1626  & 154  & 2340 \\
left axis deviation                  & 39732003  & 121 & 1545  & 0 & 940 & 5146  & 138  & 7769 \\ 
\begin{tabular}[c]{@{}l@{}}\vspace{-0.5em}left bundle\\ branch block\end{tabular}            & 164909002  & 104 & 240 & 274  & 231 & 536 & 84 & 1365 \\ 
low QRS voltages                     & 251146004 & 125 & 1043  & 0  & 374  & 182 & 322 & 1921 \\ 
\begin{tabular}[c]{@{}l@{}}\vspace{-0.5em}premature atrial\\contraction\end{tabular}         & \multicolumn{1}{l}{284470004} & \begin{tabular}[c]{@{}l@{}}\vspace{-0.5em}30\\\vspace{-0.5em} 30+308\\\vspace{-0.5em} 30+310\\\vspace{-0.5em} 30+340\\\vspace{-0.5em} 30+341\\ 30+349\end{tabular} & 1312 & 689 & 639 & 398  & 539  & 3577  \\ 
right axis deviation                 & 47665007  & 120  & 853  & 1 & 83 & 343 & 161 & 1441  \\ 
\begin{tabular}[c]{@{}l@{}}\vspace{-0.5em}right bundle \\ branch block\end{tabular}            & 59118001  & 106 & 649 & 1858  & 542& 0 & 710  & 3759  \\
sinus arrhythmia                     & 427393009  & 23  & 2550 & 11 & 455 & 772   & 1553  & 5341   \\ 
sinus bradycardia                    & 426177001  & 22   & 16559  & 45   & 1677  & 637   & 2711  & 21629  \\ 
sinus rhythm                        & 426783006  & -\tnote{2}  & 8125  & 922  & 1752  & 18172   & 16858\tnote{2} & 45829 \\ 
sinus tachycardia                    & 427084000  & 21 & 7255  & 303  & 1261 & 827  & 725  & 10371  \\ 
T wave abnormal                      & 164934002   & 147 & 7043  & 22  & 2306  & 2345  & 2042  & 13758  \\ 
T wave inversion                     & 59931005  & 147+367  & 2877  & 5 & 812 & 294 & 176 & 4164 \\ \hline
\end{tabular}
\egroup
\begin{tablenotes}
    \small
    \item[1] The prolonged PR interval label was merged to the first degree AV block label. 
    \item[2] Due to the absence of the sinus rhythm label in the SPH data, the labels were imputed.
\end{tablenotes}
\end{threeparttable}
\end{adjustbox}
\end{table}

\subsubsection{Imputation of the sinus rhythm label} 

Establishing a mapping for the sinus rhythm (SR) label between AHA statements and SNOMED CT codes required a different approach because SR is not explicitly included in AHA statements but is in SNOMED CT codes. The AHA statements clearly differentiate between "normal" ECGs and those exhibiting abnormalities. Normal ECGs are simply assumed to be in sinus rhythm without any abnormalities. However, in the case of abnormal ECGs, it remains unclear whether the recording is in sinus rhythm. Due to the complexity of diagnoses, we chose to impute the SR label when mapping abnormal ECGs from the AHA statements to the SNOMED CT codes.

The imputation of the SR label to the SPH source included the following steps: 1) A one-hot-encoded $n \times 16$ label matrix, where $n$ is the number of the ECGs and the 16 columns correspond to the labels (excluding SR), was constructed both for the CinC and SPH sources. 2) A logistic regression model ($\lambda$=0.01) was trained on the CinC data to predict the SR label, using the other 16 diagnoses as input features. 3) The logistic regression model was used to impute the SR label to ECGs of the SPH source labeled as "abnormal". The once with "normal" label were by default assigned the positive SR label. Overall, the SR label was assigned to 16,858 (72.4 \%) out of 23,274 ECG recordings in the SPH source.

\subsection{Predictive models}

For ECG classification, we used two deep neural network architectures and two traditional machine learning methods: a residual neural network, a transformer neural network, logistic regression (LR), and XGBoost. The neural network architectures include the top two performers from the PhysioNet/CinC Challenge 2020: the residual neural network is based on the second-place entry \cite{zhao_repo} and the transformer neural network on the winning entry \cite{natarajan2020wide}.

The data preprocessing was identical for both neural networks and followed similar principles as in \cite{zhao_repo}. All data were resampled to 250 Hz and each ECG was adjusted to a length of 4096 samples, approximately 16 seconds. Signals over 16 seconds were randomly cropped. For shorter signals, zero padding was applied to both sides of the signals to extend them to the required sample size, with the proportion of padding between the sides determined randomly. The signal amplitudes were normalized to the range [0,1]. Additionally, sex values were encoded using one-hot encoding, and age values were scaled to the range [0,1].

For the benchmark models, LR and XGBoost, we extracted 20 features from lead II, replicating the approach used in the transformer implementation \cite{natarajan2020wide}, which also incorporated these hand-crafted features. They employed a random forest to assess feature importance and selected the top 20 features from over 300 extracted ECG features based on the results. These features include heart rate variability features and morphological features. Additionally, age and sex features were included in the feature set, resulting in a total of 22 features that are used in classification.

\subsubsection{Improved residual neural network}

The first model \cite{zhao_repo} is a residual neural network (ResNet) architecture with additional squeeze and excitation (SE) blocks. The ResNet consists of a single convolutional layer, followed by eight residual blocks (RBs). Each of these blocks comprises two convolutional layers and the SE block. Both the first convolutional layer and the RB units employ 64 convolutional filters, with the number of filters doubling after every second RB unit. The feature dimension is reduced by half after the max pooling layer and the third, fifth, and seventh RBs. The ResNet was used with binary cross entropy loss and the Adam optimizer with an initial learning rate of 0.003.

Compared to the original ResNet architecture, the improved ResNet includes four modifications: Firstly, the age and sex information are passed to the network through an additional fully connected layer before being added to the final fully connected layer. Secondly, to enhance the learning of meaningful features, a kernel size of 15 is applied to the first convolutional kernel, while the size of the other kernels is 7. Thirdly, a dropout layer with a dropout rate of 0.2 is placed between the two convolutional kernels in each RB to reduce the risk of overfitting. Finally, a SE block is added into each RB to model the spatial relationships between the ECG channels. 

\subsubsection{Wide and deep transformer neural network}

The second model \cite{natarajan2020wide} is a wide and deep transformer neural network. The wide component incorporates the 22 hand-crafted features, while the deep component includes three modules: 1) An embedding network that applies a series of convolution operations to capture the latent space representation of the ECG signal; 2) a transformer module; and 3) a multi-label classification head with fully connected layers to generate 64 deep features. These deep features are concatenated with the wide features and fed into a final fully connected layer that produces outputs corresponding to the diagnostic labels.

Similarly to the ResNet, binary cross entropy loss was used. However, instead of the Adam optimizer, the Noam optimization was employed as in the original implementation. This approach incorporates the Adam optimizer within a scheduling routine that linearly increases the learning rate during a 4000-step warm up phase and then decreases it according to the inverse square root of the step number.

\subsubsection{Benchmark models}

For the LR classifier, the multilabel classification was performed using one-vs-the-rest strategy, where one classifier is fit for each label. The maximum number of iterations was adjusted to 2000 to ensure convergence. Aside from this, the default hyperparameter values were used. The XGBoost classifier was trained using default hyperparameter values.

\subsection{Performance evaluation}

\subsubsection{Classifier performance metrics}

We used two metrics to evaluate the model performance on the multilabel classification task: macro-averaged and micro-averaged area under the receiver operating characteristics (AUC) \cite{hanley1982AUC}. The macro-averaged AUC is computed by averaging the sum of the AUCs for each label, treating each label equally. The micro-averaged AUC is computed by first combining all the predictions and true labels and calculating the overall True Positive Rate (TPR) and False Positive Rate (FPR) to provide a ROC curve. The AUC is then computed for that ROC curve.

The metrics are computed for a filtered subset of the predictions, along with the corresponding actual labels. This means, before computation, we filtered out labels and their corresponding predictions that not present in the actual labels of the training or test sets. For example, as shown in Table \ref{table:databases_diagnoses}, the PTB and PTB-XL source do not contain any ECGs indicating right bundle branch block, and therefore, we filtered them out to facilitate the computation of the metrics.

\subsubsection{Cross-validation methods}

\begin{figure}
    \centering
    \includegraphics[width=1\linewidth]{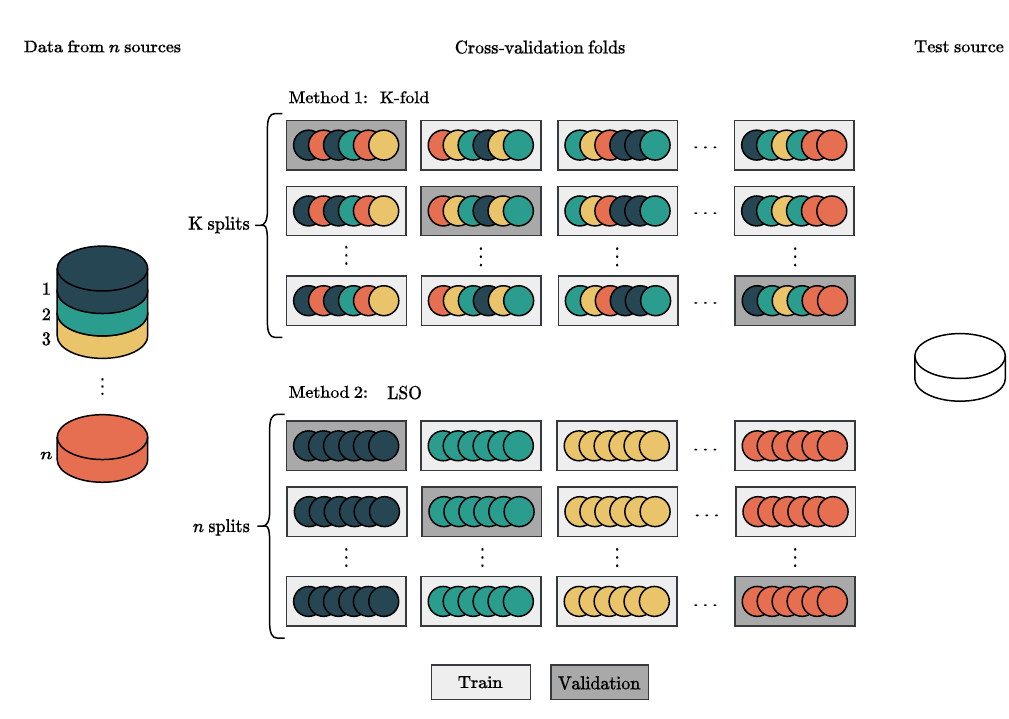}
    \caption{Illustration of the difference between the K-fold CV and the LSO CV. The training data is gathered from $n$ distinct sources. In the K-fold CV, a training set is first pooled from all the sources and then randomly divided into K disjoint folds, whereas in the LSO CV, the data sources correspond to distinct folds. Each split corresponds to one round of CV, where a model is trained on the union of train folds and used to make predictions on the validation fold. The final performance estimate is the average of the validation fold performances. The final model is trained on the full training set consisting of all the data from the $n$ sources. A test source refers to out-of-sample data that is not part of the $n$ sources used for training and cross-validation, but on which the model would later be applied on.}
    \label{fig:cv_experiments}
\end{figure}

In this study, we consider two variants of cross-validation (CV): stratified K-fold cross-validation (K-fold CV), and leave-source-out CV (LSO CV). Figure~\ref{fig:cv_experiments} illustrates the difference between the two CV methods. In CV, the dataset is repeatedly partitioned into a training set used for fitting a model, and validation set used to evaluate the prediction performance of the model. The performance estimates from different CV repetitions are averaged to obtain the final performance estimate. Compared to using a single data split, CV has the advantage of allowing the use of all the data to validate the model. CV methods have also been used derive confidence intervals for prediction performance, though this has been shown to be challenging due to dependencies between the repetitions \cite{bates2023cross}. Regardless of CV approach used, the final model is trained on all the training data combined together.

Standard K-fold CV consists of randomly partitioning the dataset to K equally sized non-overlapping folds (typically, K=10 or K=5) \cite{wong2015performance, kohavi}. On each round of CV, K-1 folds are use as the training set, and the remaining fold as the validation set. In stratified K-fold CV the folds are sampled in a way that preserves the class distribution in the original dataset. Stratification can be especially important when classes with very low membership are present in the data, to ensure that sufficient representation from each class in present in all the folds. K-fold CV can be expected to perform well if the training set can be assumed to be an i.i.d. sample from the same population on which the prediction model will later be applied on. This basic assumption is likely to be violated in applications where the goal is to make predictions on data from new sources (e.g. a hospital) not represented in the data used to develop the model. This can lead to unreliable performance estimation.
For the standard i.i.d. setting, K-fold cross-validation is known to have a negative bias, because models trained on K-1 folds tend to perform slightly worse than ones trained with the full training set, especially for small values of K. Empirically, it has been shown that with large enough K (e.g. K=10 or $20$) this issue is mostly resolved \cite{kohavi}. On the other hand, for multi-source data, K-fold can be expected to have positive bias for out-of-source prediction performance estimation, because same the sources are represented in both training and validation folds \cite{geras2013multiple}.

LSO CV \cite{rakotomalala2006accuracy,geras2013multiple} is a method designed to be applied on data arising from different sources which have different generating processes. The goal is to estimate the out-of-source prediction performance. The method works by partitioning the data so that each fold corresponds to data points coming from a single source. On each round of CV, one source is left out as validation set, and the rest are combined to form the training set. In the context of our study, each source corresponds to a single data source described e.g. in Table \ref{table:databases_overview}, and LSO CV estimates the expected prediction performance of the model for patients from a new source not part of the data used to develop and validate the model. While LSO CV does not suffer from the same positive bias as K-fold does, it can be expected to have a negative bias \cite{geras2013multiple}. This is because it estimates the prediction performance for a model trained on n-1 sources, whereas the final model trained on all the n sources is likely to perform better, especially if the number of sources is small. This negative bias cannot be reduced straightforwardly by increasing the number folds, since this would require access to new data sources.

\subsubsection{Performance metrics for comparing cross-validation methods}

To measure how well K-fold CV and LSO CV estimate prediction performance on a new source, the following metrics are employed. We perform repeated experiments on $n$ distinct data sources, each corresponding to patients from one source. Let $i\in\{1,2,...,n\}$ be the index for an out-of-source test source, $\hat{y}_i$ the CV AUC and $y_i$ the test AUC for a model trained on data from other sources. Further, let $e_i=\hat{y}_i-y_i$ denote the signed error for the CV estimate. Then, $\mu=\frac{1}{n}\sum_{i=1}^n e_i$ denotes the mean error (ME) over the experiments, $\sigma=\sqrt{\frac{1}{n-1}\sum_{i=1}^n (e_i-\mu)^2}$ the standard deviation (SD), and  $\sqrt{\mu^2+\sigma^2}$ the root mean square error (RMSE). The ME captures any systematic bias, with positive value corresponding to overoptimistic and negative to pessimistic CV results. The standard deviation measures the random component of the error, and together these form the overall RMSE. 

\subsection{Experimental setup}

We performed two sets of experiments in order to evaluate how reliable CV estimates are, when the goal is to apply the trained model for making prediction to data from a new source. In the first set of experiments we simulated the common setting where all the data for model training and validation comes from a single source. In the second set of experiments, we considered the setting where data from multiple sources is available for training and validation. 

For the K-fold CV method, we used the iterative multi-label stratification introduced in \cite{iterative-strat} to preserve nearly similar proportions of each diagnosis within the data folds. Unlike label set-based stratification, which considers different combinations of labels, iterative stratification independently considers the proportions of individual labels. Like many medical and multi-label datasets, there is an imbalance in the label proportions \cite{he2009}, and this also concerns our dataset.

The experiments were run using Python software package 3.10.4 with Torch version 1.13.1. We used the Puhti supercomputer provided by the CSC – IT Center for Science Ltd. to run and evaluate the neural network models. CSC offers high-quality information technology services to academic institutions, research institutes and businesses in Finland. The LR classifier was implemented using the scikit-learn Python library version 1.3.1 and the XGBoost classifier was implemented using xgboost Python library version 2.1.0. The code is freely available on GitHub, with the link provided in the Code Availability section. 

\subsubsection{Experimental setup for single source data}

In the single source experiments, we repeated the same procedure for each pair of sources in the data: First, one source was chosen as a training set and another as a test set. To estimate performance on the training source, the stratified 5-fold CV was conducted. Here, the performance estimate was calculated as the average over the five train/validation folds performed within the CV.

Finally, we trained a model on the full training set, and evaluated the AUC on the test set. The ME, SD, and RMSE were calculated for each training source separately, by comparing the 5-fold CV AUC to the test AUCs of the model on the other sources.

\subsubsection{Experimental setup for multi-source data}

In the multi-source experiments, one data source was left out as a test set, and the remaining four sources were used together as a training set. The experiment was repeated five times, so each source served in turn as the test set.

For the training set, we employed two CV methods: the stratified 4-fold CV, where folds consist of pooled data from all the sources, as well as the LSO CV, where data from each source corresponded to its own fold (Figure~\ref{fig:cv_experiments}). The reason for choosing the value of K as 4 is because LSO also divides the training set into 4 folds, with each fold corresponding to one data source in the training set.

The full training set was used to fit the final model and the test set was used to evaluate the performance. Thus, we ended up training five final models, one for each test set. The ME, SD, and RMSE were computed for both the CV methods over five experiments, each corresponding to a different choice of test source. The final models for both the 4-fold CV and the LSO CV were the same, as the training set always included four out of five sources, with the remaining source used as the test set.

\subsubsection{Experimental setup for source prediction experiment}

The central assumption underlying the use of LSO CV is that there are systematic differences between data coming from different sources. From Figure~\ref{fig:label_occur} and Table~\ref{table:databases_diagnoses} one can already observe the clear differences in label distributions. We further employ a heuristic based on training a multi-class classifier to evaluate the degree to which the data sources can be distinguished from each other. In this experiment, we combined the data from all the sources and divided it with 70 \% allocated for training and 30 \% for testing sets using stratification based on the sources. A ResNet classifier \cite{zhao_repo} was then trained to predict, from which of the five data sources a patient comes from. If all the data came from same distribution, the expected classification performance would be on random level, or at most equal to relative fraction of ECGs coming from the largest source for a majority voter classifier. Achieving higher classification performance can be taken as indicative of there being clear differences between the sources.

Three different input sets were used to determine which features contribute most significantly to accurate hospital identification. The input sets were 1) ECGs, sex and age, representing the initial feature set used in the subsequent multilabel classification; 2) ECG, sex, age and diagnoses, i.e. the labels in the ECG classification task; and 3) diagnoses only. The diagnoses were incorporated through the final fully-connected layer, consistent with the original architecture of the ResNet, which integrated age and sex features in this manner. Additionally, for the multiclass classification task where the outputs were the data sources $s_n, n=1,...,5$, the Softmax function was used for activation instead of the originally used Sigmoid function, and cross entropy was used for computing the loss. We visualized the outcomes in confusion matrices to identify whether some sources were more prone to misclassification and which source would commonly be misclassified as another. For classification metric, we computed classification accuracy.

\section{Results}\label{sec:results}

We carried out two distinct sets of experiments in single source and multi-source settings. These experiments were designed to provide insight on the reliability of CV estimates, when the final goal is to apply the trained model to patient populations from new data sources not represented in the training data. {Additionally, we analyzed how accurately one can distinguish between the patient cohorts across the five data sources to gain a deeper understanding of the differences and potential biases among the sources.

\subsection{Single source experiments}

The results for the experiments, where data from a single source was used for training and CV are summarized in Figure~\ref{fig:5fold_results} per training set. It's noteworthy that the models trained on either of the CinC sources result in the highest test macro- and micro-AUC for the SPH source. In some instances, the test AUCs exceed the corresponding CV estimates.

The CV results are generally higher than test result corresponding to the same training set for each model. In terms of the macro-AUC, the CPSC and CPSC-Extra source show greater variation in CV estimates compared to the other data sources, regardless of which model was trained. The ResNet achieves higher CV and test AUCs than the other models irrespective of the test source, whereas the Transformer tends to exhibit greater variation in CV estimates. A similar trend is observed for the micro-AUC, with larger variations in test AUCs: Regardless of the trained model, the CV estimates often exceed 0.9 and surpass the corresponding test AUCs. When models are trained on the CPSC and CPSC-Extra source, the micro-AUCs are also lower compared to those trained on other data sources.

\begin{figure}
    \centering
    \includegraphics[width=1\linewidth]{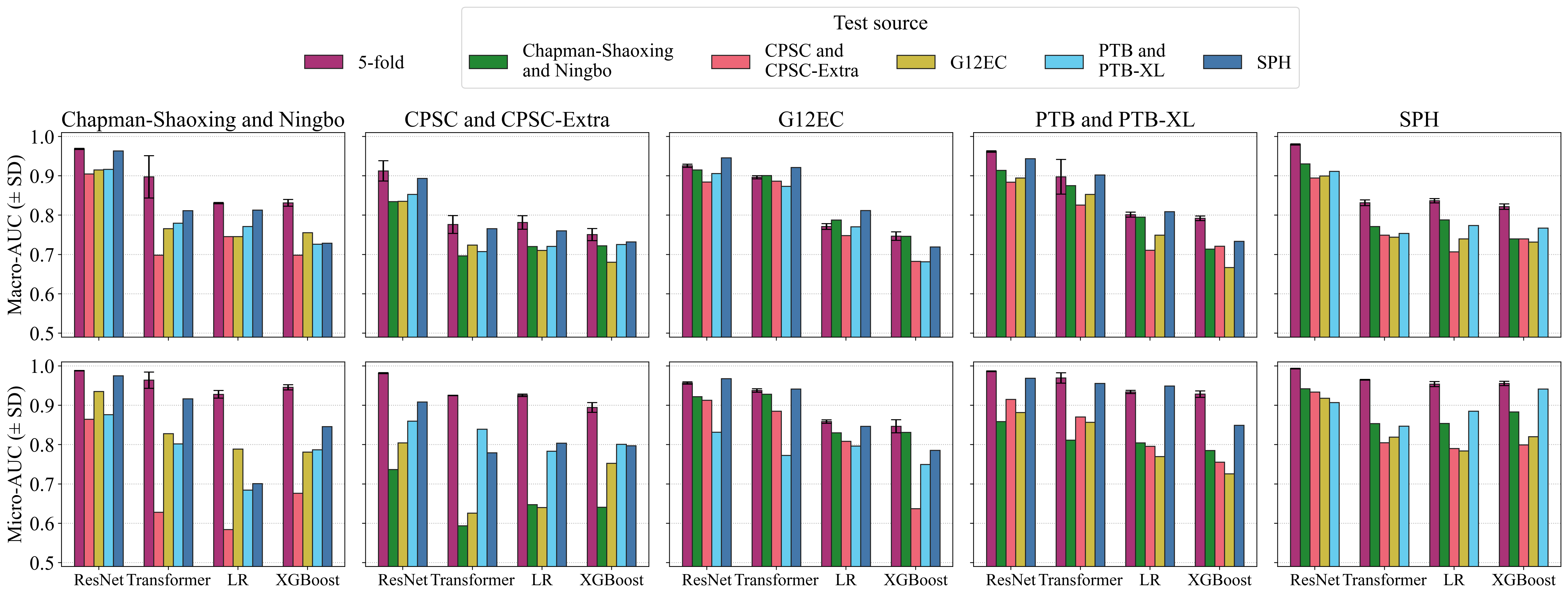}
    \caption{AUC ($\pm$ SD) scores for the 5-fold CV and individual test sets per training set}
    \label{fig:5fold_results}
\end{figure}

The ME, SD and RMSE for between CV AUCs and test AUCs are summarized in Table \ref{tab:5fold_errors} per training source. With a single exception, the MEs are positive for all the models regardless of the training source, demonstrating overoptimistic bias for 5-fold CV estimates when compared to test performance on an independent data source.

\begin{table}
\caption{Assessment of the reliability of the 5-fold CV per training set in the single-source experiment}
\label{tab:5fold_errors}
\centering
\begin{adjustbox}{width=0.8\textwidth,center=\textwidth}
\bgroup
\def\arraystretch{1.5}
\begin{tabular}{llrrrrrr}
\hline
 &  & \multicolumn{3}{c}{\textbf{Macro-AUC}} & \multicolumn{3}{c}{\textbf{Micro-AUC}} \\ \hline
\textbf{Train source} & \multicolumn{1}{c}{\textbf{Method}} & \multicolumn{1}{c}{\textbf{Mean}} & \multicolumn{1}{c}{\textbf{SD}} & \multicolumn{1}{c}{\textbf{RMSE}} & \multicolumn{1}{c}{\textbf{Mean}} & \multicolumn{1}{c}{\textbf{SD}} & \multicolumn{1}{c}{\textbf{RMSE}} \\ \hline
\multirow{4}{*}{\textbf{\begin{tabular}[c]{@{}l@{}}Chapman-Shaoxing \\ and Ningbo\end{tabular}}} 
& ResNet & 0.0434 & 0.0260 & 0.0506 & 0.0754 & 0.0518 & 0.0914 \\
 & Transformer & 0.1338 & 0.0477 & 0.1420 & 0.1703 & 0.1207 & 0.2088 \\
 &  LR & 0.0618 & 0.0318 & 0.0695 & 0.2383 & 0.0838 & 0.2526 \\
 & XGBoost & 0.1040 & 0.0234 & 0.1066 & 0.1730 & 0.0704 & 0.1868 \\  \hline
\multirow{4}{*}{\textbf{\begin{tabular}[c]{@{}l@{}}CPSC and \\ CPSC-Extra\end{tabular}}} 
& ResNet & 0.0586 & 0.0276 & 0.0648 & 0.1544 & 0.0738 & 0.1711 \\
 & Transformer & 0.0529 & 0.0304 & 0.0610 & 0.2156 & 0.1185 & 0.2460 \\
 & LR & 0.0535 & 0.0220 & 0.0579 & 0.2068 & 0.0869 & 0.2244 \\
 & XGBoost & 0.0358 & 0.0234 & 0.0428 & 0.1465 & 0.0747 & 0.1645 \\ \hline
\multirow{4}{*}{\textbf{G12EC}} & ResNet & 0.0129 & 0.0255 & 0.0286 & 0.0482 & 0.0566 & 0.0743 \\
 & Transformer & 0.0010 & 0.0205 & 0.0205 & 0.0558 & 0.0766 & 0.0947 \\
 & LR & -0.0083 & 0.0269 & 0.0282 & 0.0383 & 0.0223 & 0.0443 \\
 & XGBoost & 0.0395 & 0.0313 & 0.0504 & 0.0958 & 0.0828 & 0.1267 \\ \hline
\multirow{4}{*}{\textbf{\begin{tabular}[c]{@{}l@{}}PTB and\\  PTB-XL\end{tabular}}} & ResNet & 0.0529 & 0.0263 & 0.0591 & 0.0805 & 0.0477 & 0.0936 \\
 & Transformer & 0.0336 & 0.0326 & 0.0468 & 0.0960 & 0.0603 & 0.1133 \\
 &  LR & 0.0352 & 0.0448 & 0.0570 & 0.1041 & 0.0808 & 0.1318 \\
 & XGBoost & 0.0834 & 0.0291 & 0.0883 & 0.1496 & 0.0528 & 0.1586 \\ \hline
\multirow{4}{*}{\textbf{SPH}} &ResNet & 0.0707 & 0.0160 & 0.0725 & 0.0682 & 0.0158 & 0.0700 \\
 & Transformer & 0.0772 & 0.0119 & 0.0781 & 0.1339 & 0.0230 & 0.1358 \\
 & LR & 0.0845 & 0.0364 & 0.0920 & 0.1256 & 0.0493 & 0.1349 \\
 & XGBoost & 0.0772 & 0.0156 & 0.0788 & 0.0946 & 0.0644 & 0.1144 \\ \hline
 \multicolumn{6}{l}{\footnotesize LR = Logistic regression}
\end{tabular}
\egroup
\end{adjustbox}
\end{table}

\subsection{Multi-source experiments}

The results from the multi-source experiments, where data from multiple sources were used for training and validation are summarized in Figure~\ref{fig:multisource_results}. The results are reported per test set. The results of the 4-fold CV were higher than the LSO CV for both macro- and micro-AUC regardless of the model, as expected. The 4-fold CV AUCs are very similar regardless of the test source for each model, and variance over the folds is small. In contrast, there is more variability between LSO CV results for different test sources and also between the folds. The LSO CV estimates for both macro- and micro-AUC tend to be closer to the test AUCs than the 4-fold CV estimates. The ResNet and Transformer networks consistently achieve higher AUCs for each test source compared to LR and XGBoost, particularly in the LSO CV.

\begin{figure}
    \centering
    \includegraphics[width=1\linewidth]{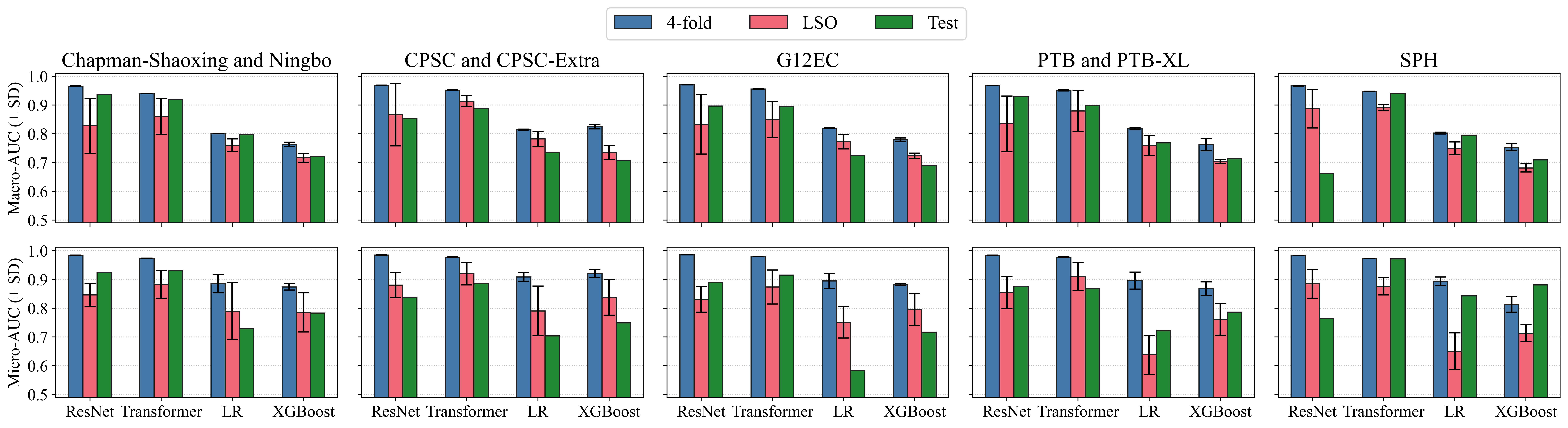}
    \caption{AUC ($\pm \text{SD}$) scores for the 4-fold CV, the LSO CV and the test results per test set}
    \label{fig:multisource_results}
\end{figure}

The comparison between the errors of 4-fold CV and LSO CV estimates is presented in Table \ref{table:errors} for each model. The mean errors for the LSO CV are close to zero, indicating unbiased performance estimates In contrast, the 4-fold CV results mean errors demonstrate clear optimistic bias. On the other hand, the LSO CV has higher SD for the errors. Combining these two components of error together, LSO CV has lower RMSE, providing overall more reliable performance estimates in the experiments than 4-fold CV.

\begin{table}[]
\caption{Assessment of the reliability of the CV methods in the multi-source experiment} \label{table:errors}
\centering
\begin{adjustbox}{width=0.8\textwidth,center=\textwidth}
\bgroup
\def\arraystretch{1.5}
\begin{tabular}{clrrrrrrrr}
\hline
\multicolumn{1}{l}{} &  & \multicolumn{2}{c}{\textbf{ResNet}} & \multicolumn{2}{c}{\textbf{Transformer}} & \multicolumn{2}{c}{\textbf{LR}} & \multicolumn{2}{c}{\textbf{XGBoost}} \\ \hline
\multicolumn{1}{l}{} &  & \multicolumn{1}{c}{\textbf{4-fold}} & \multicolumn{1}{c}{\textbf{LSO}} & \multicolumn{1}{c}{\textbf{4-fold}} & \multicolumn{1}{c}{\textbf{LSO}} & \multicolumn{1}{c}{\textbf{4-fold}} & \multicolumn{1}{c}{\textbf{LSO}} & \multicolumn{1}{c}{\textbf{4-fold}} & \multicolumn{1}{c}{\textbf{LSO}} \\ \hline
\multirow{3}{*}{\textbf{\begin{tabular}[c]{c}Macro-\\ AUC\end{tabular}}} & Mean & 0.1121 & -0.0061 & 0.0403 & -0.0299 & 0.0468 & 0.0005 & 0.0684 & 0.0043 \\
 &  SD & 0.1128 & 0.1372 & 0.0256 & 0.0336 & 0.0409 & 0.0446 & 0.0330 & 0.0260 \\
 &  RMSE & 0.1590 & 0.1373 & 0.0477 & 0.0450 & 0.0622 & 0.0446 & 0.0759 & 0.0264 \\ \hline
\multirow{3}{*}{\textbf{\begin{tabular}[c]{c}Micro-\\ AUC\end{tabular}}} & Mean & 0.1261 & 0.0012 & 0.0621 & -0.0215 & 0.1800 & 0.0084 & 0.0884 & -0.0049 \\
 & SD & 0.0605 & 0.0812 & 0.0427 & 0.0583 & 0.0937 & 0.1441 & 0.0963 & 0.1033 \\
 & RMSE & 0.1399 & 0.0812 & 0.0754 & 0.0621 & 0.2029 & 0.1443 & 0.1307 & 0.1034 \\ \hline
\multicolumn{6}{l}{\footnotesize LR = Logistic regression}
\end{tabular}
\egroup
\end{adjustbox}
\end{table}

\subsection{Source prediction experiment}

The confusion matrices per input set are visualized in Figure~\ref{fig:confusion_matrices}. The corresponding accuracies for the input sets were 1) 0.969, 2) 0.973, and 3) 0.625. A trivial baseline model that always predicts the largest class (Chapman-Shaoxing and Ningbo) achieves 0.424 accuracy. Clearly, the model that uses only diagnoses already performs higher than this leveraging systematic differences in label distributions. However, much higher performance still can be achieved when making predictions based on ECG, age and sex. Incorporating diagnoses as additional features to this model did not lead to noticeable improvement in accuracy.

\begin{figure}
    \centering
    \includegraphics[width=\linewidth]{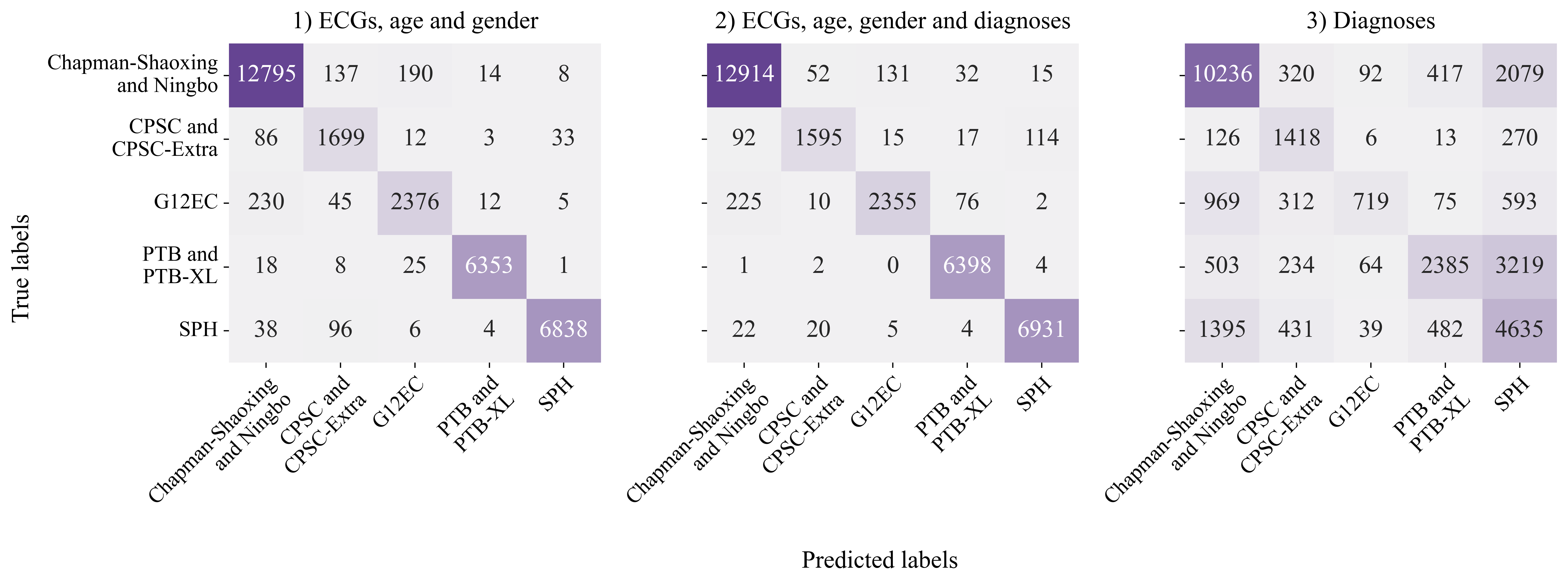}
    \caption{Confusion matrices for each input set in the multiclass classification task where the output is the data source of ECGs}
    \label{fig:confusion_matrices}
\end{figure}

The confusion matrices indicate that the Chapman-Shaoxing and Ningbo source was more frequently misclassified, and other sources were often misclassified as this source. This occurs especially when using only diagnoses as an input: The ECGs from the G12EC source are more often misclassified as originating from the Chapman-Shaoxing and Ningbo source than they are correctly classified. Similarly, ECGs from the PTB and PTB-XL source is often misclassified as the SPH data source.

\section{Discussion}\label{sec:discussion}

The experiments comparing stratified K-fold CV and LSO CV revealed that models trained in one context may exhibit poor performance on a new data source. While this issue is well-known in research \cite{geras2013multiple, rakotomalala2006accuracy,padovano2022hidden, white2018}, it has not been extensively analyzed within medical data. A major contributing factor behind the issue has been the lack of sufficiently diverse datasets \cite{Jiang230,kulkarni2021}. However, even with the availability of data, its characteristics play a significant role in model performance and generalizability. Within medical data, there is considerable variability in patient demographics, clinical settings, and data collection methods, all of which can influence how well a model performs on new context and highlights the importance of evaluating the robustness of predictive models — whether based on deep learning or classical techniques — across diverse datasets. Thus, it is essential not only to apply CV methods adequately but also to account for the data's characteristics and variability, as these factors influence the effectiveness of the methods. This is especially important to avoid over-optimistic performance claims (e.g., \cite{kapoor2023}). Although training highly generalizable models may not always be necessary in every scenario \cite{futoma2020}, understanding the differences between the training context and a test site remains vital for reliable model performance evaluation and deployment in healthcare applications.

Building on these challenges, issues such as varying labeling standards, data bias, and ambiguous ground truth are evident \cite{Rajpurkar2022, han2020survey, kulkarni2021, norori2021addressing}, and they remain relevant to our study. When harmonizing data from multiple sources, the inconsistencies can significantly impact model performance, and instead of learning physiological patterns in data, models may inadvertently learn the noise specific to each source. For example, we wanted to include the SR label for the SPH data as such label was included to the CinC data and otherwise, the SPH data would only contain abnormal ECGs. Although we verified the label mapping between SNOMED CT and AHA with the cardiologists, differences in labeling practices between the sources are expected: As PhysioNet also noted, they did not correct for variations in label practices - whether machine-generated or cardiologist-provided - within their data \cite{cinc2021_longer}. Our experiments using single-source data demonstrated that models trained on either of the CinC data sources performed well with the SPH data, yet displayed increased uncertainty when applied to other CinC sources, which may suggest potential biases and differences between the data sources. Additionally, variation in data source sizes (Table \ref{table:databases_overview}) and label prevalence (Table \ref{table:databases_diagnoses} and Figure \ref{fig:label_occur}) put more emphasis on certain labels, which may skew model generalization and is often observed in medical data. Of the data sources, the CPSC and CPSC-Extra source exhibit a high degree of variety in label proportions and showed low test performance especially in single source experiments. Moreover, labeling is not the only factor distinguishing the data sources; age and sex distributions (Table~\ref{table:demo_info}), as well as recording conditions and instrumentation, also contribute to differences. When models are trained on data from one source, they may inadvertently learn patterns or noise specific to that source, rather than capturing generalizable physiological patterns from ECGs.

Such multi-source experiments are only possible if multiple data sources are available and they may give a reliable picture on the performance range where the model on unseen data would lead. The main benefit in such multi-source experiment is that they provide good insights about the uncertainty levels of the prediction performance particularly when the objective is to develop models for application to new sources, such as hospitals where training data may not be available. In such context, the LSO CV can be particularly effective in providing a clearer picture of the model's actual performance on unseen data. In the development of predictive and decision-support models to healthcare, it's crucial to understand how properties of the data used to train a model can easily lead to overoptimistic expectations of the model's actual performance and thus, potentially leading to a failure in deployment. However, to further understand this effect that's a result of different training and test contexts, domain adaptation and transfer learning could be considered in the future studies.

One of the main limitations in our study is that most of the data sources -- the CPSC and CPSC-Extra source, the Chapman-Shaoxing and Ningbo source and the SPH data -- are gathered from one country, namely China, covering over 70 \% of the ECGs recordings. Despite the significance of these sources, the geographical centralization may introduce bias and limit the generalizability of models to hospitals from other countries. Within medical data, specific concerns have raised regarding the use of non-diverse and non-representative data, particularly when it inadequately represents the diversity of population, which can result in biased algorithms and, consequently, less accurate performance \cite{Arora2023}. Although we were able to identify and address duplicate recordings in the SPH source, we could not do the same for the CinC data. Consequently, duplicate entries were retained in the final dataset, as these duplicates constitute only a very small fraction of the total data and are unlikely to have a significant impact on the results. However, this is a challenge that must be acknowledged when working with medical datasets, as it can affect claims regarding model performance \cite{kapoor2023}.

Additionally, we did not evaluate different augmentations of ECGs nor tuned hyper-parameters for the neural networks, which are clear limitations for our study. Data augmentation in ECG classification has demonstrated potential to tackle with label imbalance, e.g. in forms of an amplitude alternation \cite{Do2022} and generative adversarial networks \cite{Shaker2020}. Also, albeit the success of the convolutional neural networks like ResNets for time series classification is evident, other network architectures such as a long short-term memory networks have shown their potential \cite{Zahra2020, Hiriyannaiah2021}. These are some additional factors to consider in the future studies to further develop ECG classifiers and study their capabilities in CVD classification.

While cross-validation methods are valuable tools to evaluate developed models, the context to which they are employed should be carefully considered. Researchers should stay vigilant in assessing whether the chosen approach genuinely addresses the specific research question at hand. The medical data is inherently complex, with variability stemming from differences in patient populations, data collection practices, and clinical environments, which can add challenges in model evaluation. As we demonstrated, the estimated performance of a model can vary significantly between different CV methods, albeit each method addresses different aspects of the model's generalization capabilities. For example, the K-fold CV is suitable when multiple hospitals collaborate to develop a model for use within those same hospitals. However, when the goal is to develop a model for a hospital from which training data is no available, the LSO CV may provide more reliable performance estimates. A limitation of LSO CV is that if the number of sources is small, the performance estimates obtained can be pessimistic. For example, if there is data from only two or three hospitals, removal of one of these could significantly reduce the variety of training data and hence accuracy of trained models. Yet, as also highlighted by previous studies underscoring the importance of external validation on diverse yet relevant patient populations \cite{bleeker2003external}, it can be argued that it is better to err on the side of caution than provide overoptimistic evaluations when evaluating clinical prediction models.

Regardless of the CV method, it is advisable to assess the presence of batch effects in the data as a standard practice, as demonstrated by our approach using source-based classification. It is crucial to avoid drawing misleading conclusions about a model's actual abilities in order to advance the deployment of machine learning models tailored for modern healthcare applications. In addition to performance evaluation, the LSO CV can also be used for model selection in order to choose hyperparameters, features or classifier algorithms likely to generalize well to new sources. If both model selection and final evaluation are to be performed simultaneously, nested \cite{varma2006bias} LSO CV could be implemented, though at considerable computational cost.

\section{Conclusion}\label{sec:conclusion}

In this paper we evaluated the reliability of the estimates of two CV methods, the standard K-fold CV and the LSO CV. To examine the generalization of ECG-based classifiers in multi-source context, we created a large multi-source resource by combining and harmonizing two openly available but separate datasets, the PhysioNet/CinC challenge 2021 training data and the Shandong Provincial Hospital database. We conducted a source-based classification to analyze potential biases between the data sources. For CV, we performed two sets of experiments in single and multi-source settings, a "source" corresponding to different ECG datasets. We showed that the K-fold CV provides overoptimistic performance claims of the model's generalizability power in both settings. For future studies, the enhancement of deep learning model's generalization can be examined through the utilization of data augmentation techniques and other model architectures. Additionally, to further study the effect when a model trained in one context performs poorly in another context, data adaptation and transfer learning could be considered.

\section*{Code availability}

The source code that was used to perform the experiments can be found at \url{https://github.com/UTU-Health-Research/dl-ecg-classifier}. It contains functionalities to 1) save demographic information, diagnostic labels and ECG paths to a CSV file which is used during model training and evaluation, 2) store parameters and other training and validation settings for each run as YAML files, 3) map the AHA codes to the SNOMED CT codes and vice versa, 4) preprocess ECGs using different transformations, 5) and train and run models.

\section*{CRediT authorship contribution statement}
\textbf{Tuija Leinonen}: Software, Data Curation, Formal analysis, Investigation, Writing - Original Draft, Visualization.
\textbf{David Wong}: Methodology, Writing - Review \& Editing. 
\textbf{Antti Vasankari}: Software \& Writing - Original Draft. 
\textbf{Ali Wahab}: Resources, Data Curation. 
\textbf{Ramesh Nadarajah}: Resources, Data Curation. 
\textbf{Matti Kaisti}: Supervision, Conceptualization, Methodology, Formal analysis, Writing - Review \& Editing, Funding acquisition.
\textbf{Antti Airola}: Supervision, Conceptualization, Methodology, Formal analysis, Writing - Review \& Editing, Funding acquisition.

\section*{Declaration of competing interest}
The authors declare that they have no known competing financial interests or personal relationships that could have appeared to influence the work reported in this paper.

\section*{Acknowledgments}

The authors wish to acknowledge CSC – IT Center for Science, Finland, for generous computational resources. This work has been supported by Research Council of Finland (grants 352893, 358868, 345805, 340140).




\bibliographystyle{elsarticle-num} 
\bibliography{main}
 
\end{document}